\documentclass[conference]{IEEEtran}
\IEEEoverridecommandlockouts
\usepackage{cite}
\usepackage{amsmath,amssymb,amsfonts}
\usepackage{algorithmic}
\usepackage{graphicx}
\usepackage{textcomp}
\usepackage{float}
\usepackage{xcolor}
\def\BibTeX{{\rm B\kern-.05em{\sc i\kern-.025em b}\kern-.08em
    T\kern-.1667em\lower.7ex\hbox{E}\kern-.125emX}}
\begin{document}

\title{Deepwound: Automated Postoperative Wound Assessment and Surgical Site Surveillance through Convolutional Neural Networks\\
}

\author{\IEEEauthorblockN{Varun Shenoy}
\IEEEauthorblockA{\textit{Cupertino High School} \\
Cupertino, CA, USA \\
varun.inquiry@gmail.com}
\and
\IEEEauthorblockN{Elizabeth Foster}
\IEEEauthorblockA{\textit{Castilleja School} \\
Palo Alto, CA, USA \\
ecfoster@gmail.com}
\and
\IEEEauthorblockN{Lauren Aalami}
\IEEEauthorblockA{\textit{Stanford University} \\
Stanford, CA, USA \\
laalami@stanford.edu}
\and
\IEEEauthorblockN{Bakar Majeed}
\IEEEauthorblockA{\textit{Stanford University} \\
Stanford, CA, USA \\
abbakar@stanford.edu}
\and
\IEEEauthorblockN{Oliver Aalami}
\IEEEauthorblockA{\textit{Stanford University} \\
Stanford, CA, USA \\
aalami@stanford.edu}
}
\maketitle

\begin{abstract}
Postoperative wound complications are a significant cause of expense for hospitals, doctors, and patients. Hence, an effective method to diagnose the onset of wound complications is strongly desired. Algorithmically classifying wound images is a difficult task due to the variability in the appearance of wound sites. Convolutional neural networks (CNNs), a subgroup of artificial neural networks that have shown great promise in analyzing visual imagery, can be leveraged to categorize surgical wounds. We present a multi-label CNN ensemble, Deepwound, trained to classify wound images using only image pixels and corresponding labels as inputs. Our final computational model can accurately identify the presence of nine labels: drainage, fibrinous exudate, granulation tissue, surgical site infection, open wound, staples, steri strips, and sutures. Our model achieves receiver operating curve (ROC) area under curve (AUC) scores, sensitivity, specificity, and F1 scores superior to prior work in this area. Smartphones provide a means to deliver accessible wound care due to their increasing ubiquity. Paired with deep neural networks, they offer the capability to provide clinical insight to assist surgeons during postoperative care. We also present a mobile application frontend to Deepwound that assists patients in tracking their wound and surgical recovery from the comfort of their home.
\end{abstract}

\begin{IEEEkeywords}
wound care, machine learning, mHealth
\end{IEEEkeywords}

\section{Introduction}
A critical issue in the healthcare industry, particularly in the United States, is the effective management of postoperative wounds. The World Health Organization estimates 359.5 million surgical operations were performed in 2012, displaying an increase of 38\% over the preceding eight years \cite{b1}. Surgeries expose patients to an array of possible afflictions in the surgical site. Surgical site infection (SSI) is an expensive healthcare-associated infection. The difference between the mean unadjusted costs for patients with and without SSI is approximately \$21,000 \cite{b2}. Thus, individual SSIs have a significant financial impact on healthcare providers, patients and insurers. SSIs occur in 2-5 percent of patients undergoing inpatient surgery in the U.S., resulting in approximately 160,000 to 300,000 SSIs each year in the United States alone, as summarized by \cite{b3}.

Currently, most wound findings are documented via visual assessment by surgeons. Patients revisit their surgeon a few days after the operation for this checkup. This takes up valuable time that a surgeon could use to help out other patients. Infections can also set in earlier and the delay until the checkup can exacerbate the issue. Moreover, there is a lack of quantification of surgical wounds. An automated analysis of a wound image can provide a complementary opinion and draw the attention of a surgeon to particular issues detected in a wound. Thus, a rapid and portable computer aided diagnosis (CAD) tool for wound assessment will greatly assist surgeons in determining the status of a wound in a timely manner.

Advances in software and hardware, in the form of powerful algorithms and computing units, have allowed for deep learning algorithms to solve a wide variety of tasks which were previously deemed difficult for computers to tackle. Challenging problems such as playing strategic games like Go \cite{b4} and poker \cite{b5}, and visual object recognition \cite{b6} are now possible using modern compute environments. A type of artificial neural network, called a convolutional neural network (CNN), has demonstrated capabilities for highly accurate image classification after being trained on a large dataset of samples \cite{b7}. In the past decade, research efforts have led to impressive results on medical tasks, such as automated skin lesion inspection \cite{b8} and X-ray based pneumonia identification \cite{b9}.

In this paper, we propose a novel approach to identifying the onset of wound ailments through the simple means of a picture. We introduce a CNN architecture, WoundNet, and train it with a HIPAA compliant dataset of wound images collected by patients and doctors using smartphones. Finally, we build a mobile application for the iOS software ecosystem that presents a user implementation of our CAD system. It includes clinically relevant features, such as the daily documentation of patient health and generation of wound assessments. The app enables patients to generate wound analysis reports and send them to the surgeon on a regular basis from a remote location, such as their home.

\section{Prior Work}
While the applications of machine learning in healthcare are numerous, few have attempted to solve the problem of postoperative wound analysis and surgical site monitoring. We would like to summarize two key pieces of research that sought to build models similar to the one presented in this paper.

Wang et al. showcased a comprehensive pipeline for wound analysis, from wound segmentation to infection scoring and healing prediction \cite{b10}. For binary infection classification, they obtained a F1 score of 0.348 and accuracy of 95.7\% with a Kernel Support Vector Machine (SVM) trained on CNN generated features. Their dataset consisted of 2,700 images with 150 cases positive for SSI.

Another paper by Sanger et al. used classical machine learning to predict the onset of SSI in a wound. It is trained on baseline risk factors (BRF), such as pre-operative labs (e.g. blood tests), type of operation, and a multitude of other features \cite{b11}. Their best classifier achieved a sensitivity of 0.8, specificity of 0.64, and receiver operating characteristic (ROC) area-under-curve (AUC) of 0.76. By computing the harmonic mean of their sensitivity and specificity, we determine that their F1 score is 0.71.

In our opinion, while the infection scoring model presented by Wang et al. does achieve an accuracy of 95.7\%, we believe that this metric is insufficient due to the severe class imbalance in their dataset. Sensitivity, specificity, F1 score, and ROC curves are better metrics which address this issue. This work improves upon these metrics compared to the presentation by Wang et al. While Sanger et al. have built a predictive methodology based on BRF, our approach leverages pixel data from wound images. Thus, our research complements any analysis using BRF.

According to our literature search, no prior work in dressing identification and other ailments apart from SSIs have been modeled using computational techniques. Thus, we believe we have built the most robust and comprehensive wound classification algorithm up-to-date.

\section{Materials and Methods}
\subsection{Data Collection and Description}

Prior to this research, a dataset of 1,335 smartphone wound images was collected primarily from patients and surgeons at the Palo Alto VA Hospital and the Washington University Medical Center in St. Louis. The dataset also includes images from searching the internet to counteract class imbalance. All images were anonymized and cropped into identical squares.

\begin{figure}[H]
\centerline{\includegraphics[width=0.4\textwidth]{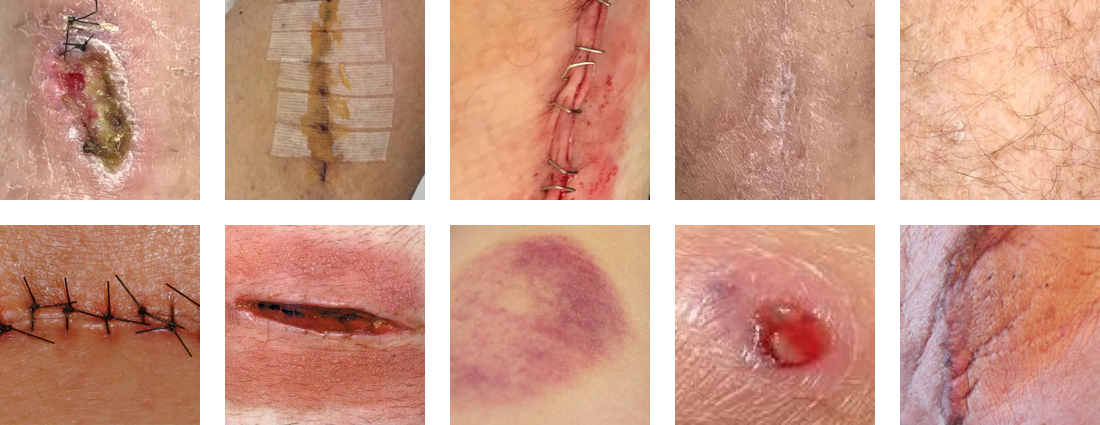}}
\caption{Example Images from the Wound Smartphone Image Dataset.}
\label{fig}
\end{figure}

Figure 1 shows a few examples of images from the dataset. As can be seen, images are very diverse and contain high variability. Images ranged from open wounds with infections to closed wounds with sutures. Table I shows the breakdown of the entire dataset.

\begin{table}[H]
\caption{Our smartphone-image wound dataset with 1,335 samples.}
\begin{center}
\begin{tabular}{|l|c|c|}
\hline
\textbf{Labels}    & \textbf{Positive} & \textbf{Negative} \\ \hline
Wound              & 615               & 720               \\ \hline
Infection (SSI)    & 355               & 980               \\ \hline
Granulation Tissue & 449               & 886               \\ \hline
Fibrinous Exudate  & 398               & 937               \\ \hline
Open Wound         & 631               & 704               \\ \hline
Drainage           & 448               & 887               \\ \hline
Steri Strips       & 129               & 1206              \\ \hline
Staples            & 98                & 1237              \\ \hline
Sutures            & 160               & 1175              \\ \hline
\end{tabular}
\label{tab1}
\end{center}
\end{table}

\subsection{Materials}
Many tools went into the development of this research. Our CNNs were engineered using the Keras deep learning framework \cite{b12} in the Python 3.7 programming language. The neural networks were trained on a Nvidia Tesla K80 GPU hosted by the Amazon Web Services Elastic Cloud Compute platform. The OpenCV computer vision library \cite{b13} was used for histogram equalization and image inspection. Scikit-learn \cite{b14} was leveraged for its variety of built-in metrics for model evaluation. The final model was deployed on a server using Flask.

A standard iOS development setup was used for the mobile application. The app was built using the Swift 4 programming language and Xcode integrated development environment.

Figure 2 below summarizes the four steps in the development of our model. We now cover each of these steps in detail.

\begin{figure}[H]
\centerline{\includegraphics[width=0.45\textwidth]{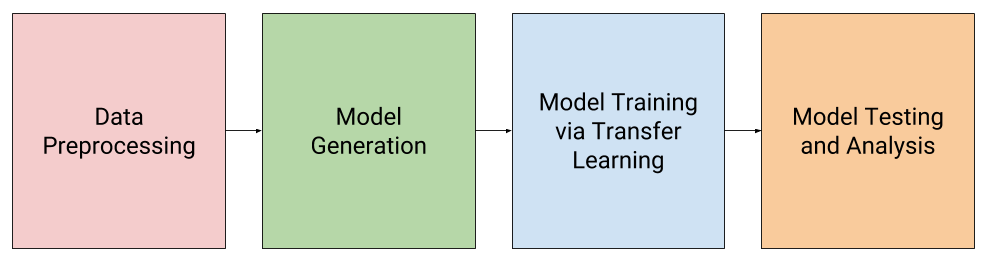}}
\caption{The pipeline for the development of Deepwound.}
\label{fig}
\end{figure}

\setcounter{figure}{4}
\begin{figure*}

 \center

  \includegraphics[width=\textwidth]{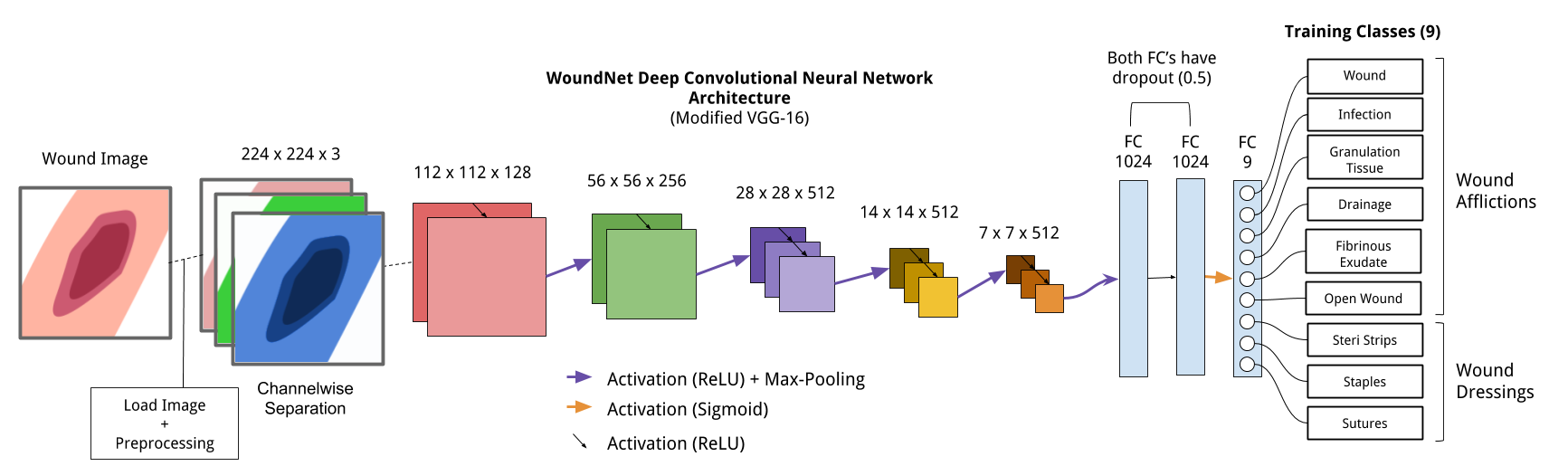}

  \caption{The WoundNet CNN Architecture.}

  \label{AAA}

\end{figure*}

In this section, we will cover the first three blocks of this pipeline. The “Model Testing and Analysis” block will be covered in the Results and Discussion section of this paper.

\subsection{Data Preprocessing}
Figure 3 gives a summary of the data preprocessing steps. Once data is loaded into memory, images are resized to 224 by 224 pixels to fit the input of our CNN architecture. The input layer is 224 by 224 by 3 pixels, the final dimension accounting for the three-color channels. We then partition the dataset into training and validation sets. 80\% of the data is used as the training data for our model and 20\% is left for model evaluation and testing.

\setcounter{figure}{2}
\begin{figure}[H]
\centerline{\includegraphics[width=0.45\textwidth]{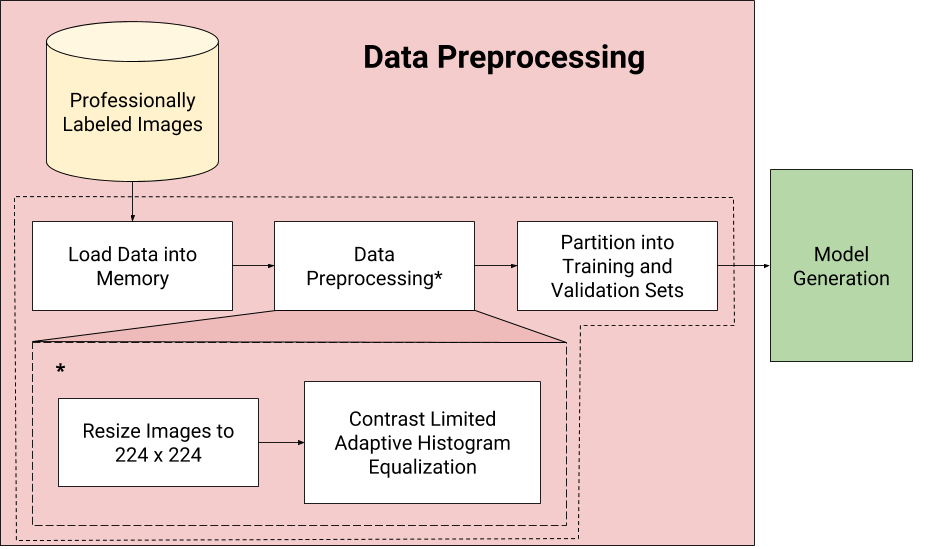}}
\caption{Prior to generating and training any CNNs, we preprocess our images using histogram equalization and image scaling.}
\label{fig}
\end{figure}

A critical component of the preprocessing stage is to compensate for vast differences in lighting and position found in smartphone images. To accommodate this, we apply contrast limited adaptive histogram equalization (CLAHE) to each image \cite{b15}. Histogram equalization (HE) takes in a low contrast image and increases the contrast between the image’s relative highs and lows to bring out subtle differences in shade and create a higher contrast image. CLAHE applies HE in individual 8x8 pixel tiles around the image. Contrast limiting is used to prevent noise from being amplified. We use a contrast limiting factor of 1. Finally, bilinear interpolation is applied to the image to remove artifacts in the borders.

\subsection{Model Generation}
The second step in the development of our model is known as model generation and is shown in Figure 4. We take the preprocessed images and generate three slightly different CNNs using the WoundNet architecture. We also prepare it for transfer learning by initializing the CNNs on ImageNet weights.

\setcounter{figure}{3}
\begin{figure}[H]
\centerline{\includegraphics[width=0.55\textwidth]{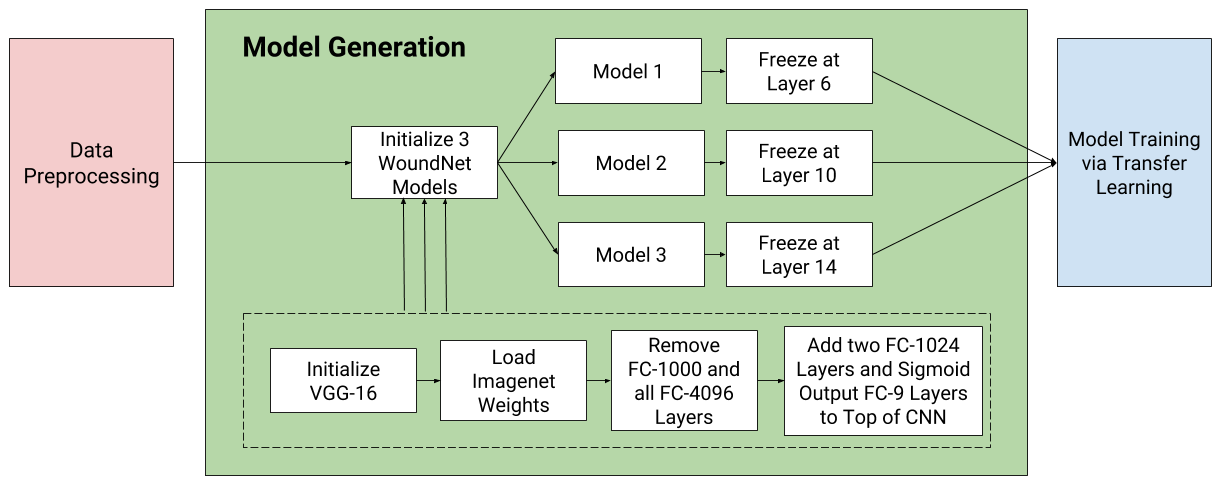}}
\caption{We generate three separate CNNs based off the VGG-16 CNN architecture and modify them for our use case.}
\label{fig}
\end{figure}
\setcounter{figure}{5}

Convolutional neural networks vastly outperform other current machine learning models for large scale image processing and classification. We make adjustments to a current state of the art CNN, VGG-16, to better suit our specific problem. The resulting configured model is known as WoundNet, illustrated in Figure 5. Three models were initialized using the VGG-16 CNN architecture \cite{b16}. While we did try other deeper network architectures, we found them to overfit on the data almost immediately, unlike VGG-16. We believe that this is due to the combination of imbalance and small size of our dataset. For example, some deeper networks   Rather than creating nine individual binary classifiers, we train each neural network to label images with all nine classes. This enables our model to find inter-label correlations through shared knowledge in the deep learning model.

We first remove the final output layer along with the two 4096-neuron fully connected (FC) layers prior to it. We append two smaller 1024-neuron FC layers, each with a dropout of 0.5, and an output layer with the sigmoid activation function. Dropout is a form of regularization that forces a chosen percentage of elements in a layer to not activate and thus reducing the overfitting of the model.

Our motivation for these changes are the following. Each wound image can be positive or negative for nine different classes in comparison to the ImageNet dataset in which every image only has a single label. Furthermore, the sigmoid function treats every class as a binary decision while softmax converts the weights of the neural network in the layer prior to the output to probabilities that add up to 1. We determined two 1024 element layers are the best choice to replace the FC layers to keep the model lightweight, faster to train, and reduce the computational complexity present in the original VGG-16 architecture.

\subsection{Model Training}
After generating our models, we fine-tune them on our data. The entire process is outlined in Figure 6. Our training phase can be divided into four critical components: transfer learning, data augmentation, training specifics, and ensembling multiple models. We will now cover each of these pieces in depth.

\begin{figure}[H]
\centerline{\includegraphics[width=0.50\textwidth]{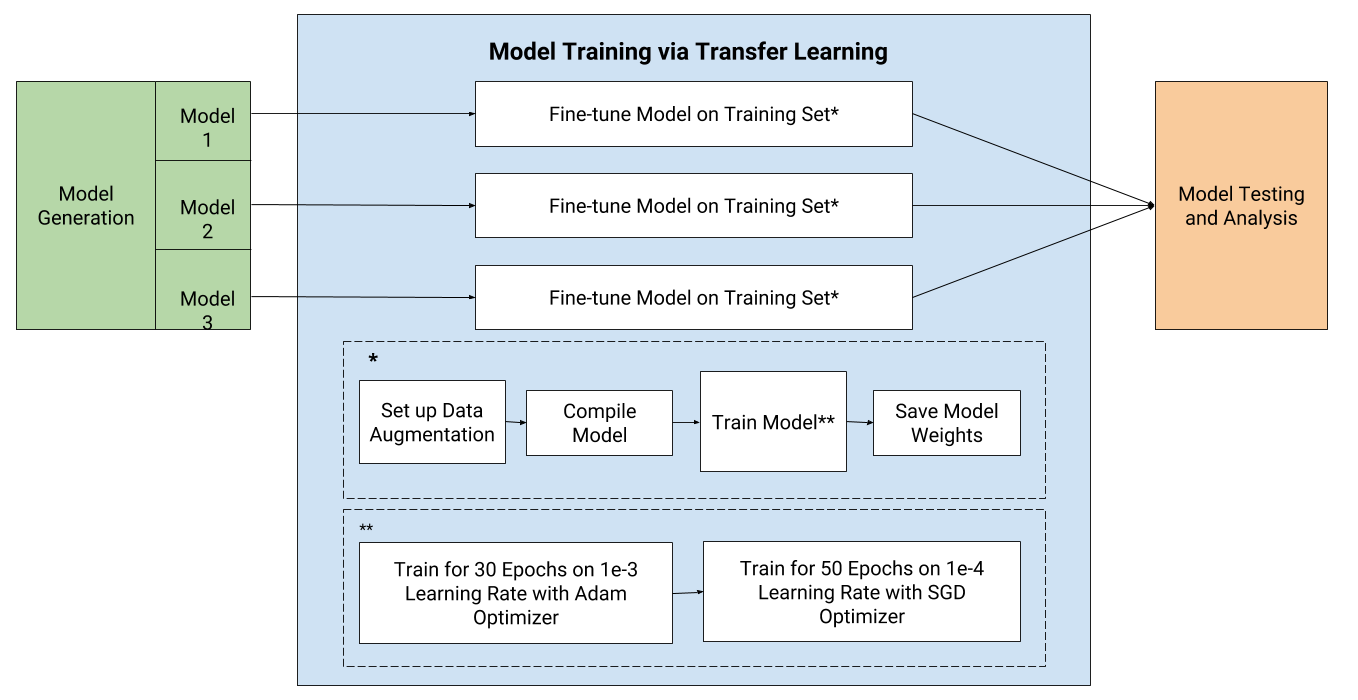}}
\caption{Model training via transfer learning.}
\label{fig}
\end{figure}

\subsubsection{Transfer Learning}
In practice, it is very difficult to train a CNN from end-to-end starting with randomly initialized weights. Furthermore, huge datasets with upwards of a million images are necessary to successfully train an accurate neural network from scratch. Too little data, such as our case, would cause a model to overfit.

We employ transfer learning \cite{b17}, also known as fine tuning, to leverage pre-learned features from the ImageNet database. The original VGG-16 model was trained from end to end using approximately 1.3 million images (1000 object classes) from the 2014 ImageNet Large Scale Visual Recognition Challenge. We use the weights and layers from the original VGG-16 model as a starting point.

Transfer learning leverages the previously learned low level features (such as lines, edges and curves). Since these features are common for any image classification task, transfer learning requires less data to arrive at a satisfactory CNN. For optimum generalization and to prevent overfitting, we freeze each of the three WoundNet models at layer 6, 10, and 14, respectively. This prevents the low-level features from being washed away from training the CNNs on the training set of wound images.

\subsubsection{Data Augmentation}
In order to make the most out of our training set, we utilize aggressive data augmentation prior to feeding images into the CNN. Data augmentation improves the generalization and performance of a deep neural network by copying images in the training set and performing a variety of random transformations on them. Each copy is randomly rotated from 0° to 360°, shifted by 10 pixels in any direction, zoomed into by a factor of 30\%, and sheared by a factor of 20\%. Copies are flipped vertically or horizontally with equal probability of 50\%. The copied data is given the same labeling as its original and is added to the current training batch. Some images generated via data augmentation are shown in Figure 7.

\begin{figure}[H]
\centerline{\includegraphics[width=0.45\textwidth]{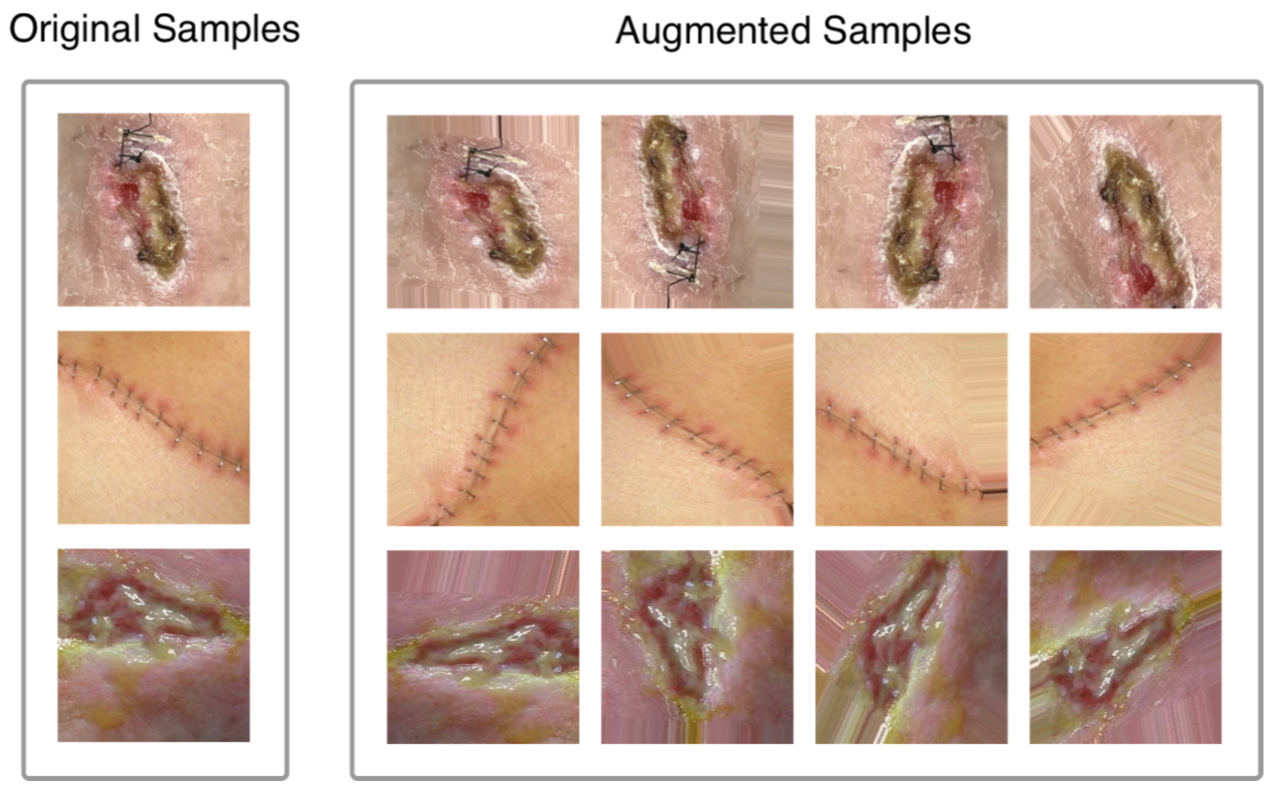}}
\caption{Examples of data augmentation.}
\label{fig}
\end{figure}

\begin{figure*}

 \center

  \includegraphics[width=\textwidth]{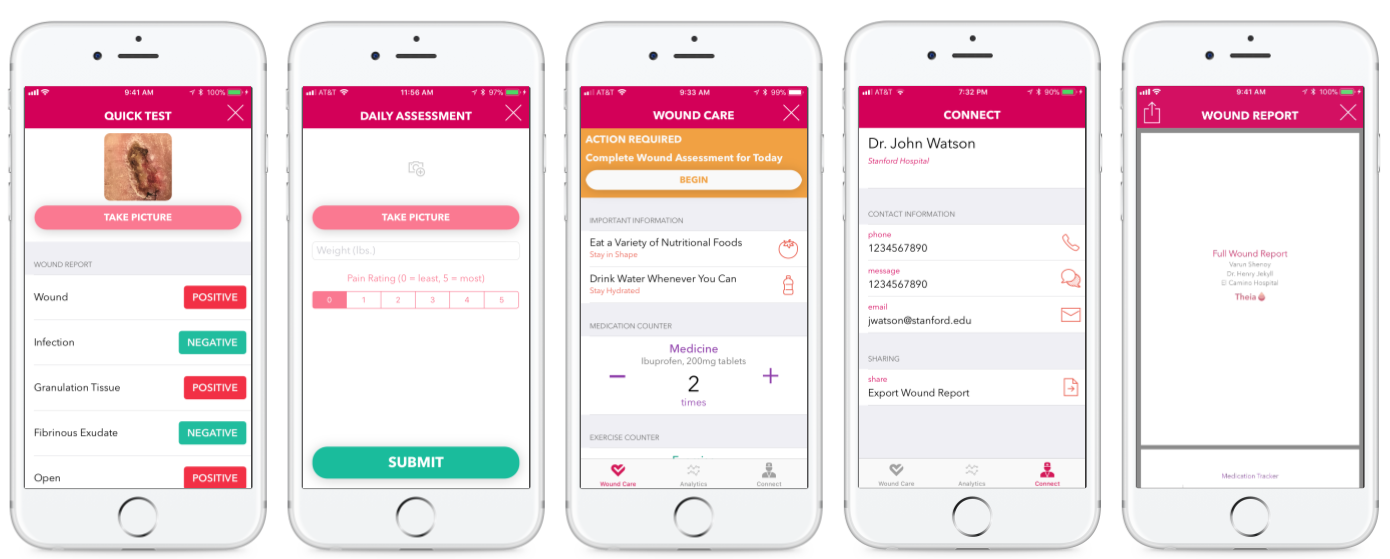}

  \caption{Screenshots from our working prototype for Theia.}

  \label{AAA}

\end{figure*}

\subsubsection{Training Specifics}
Our CNNs are trained using the backpropagation algorithm with a batch size of 64. Backpropagation is an application of the chain rule of calculus to compute loss gradients for all weights in the network. Once an image is passed through the CNN during the training phase, the error is calculated using a loss function. That loss gradient is propagated backwards through the CNN, adjusting weights in the CNN. This way, the next time the CNN sees the same image, it will arrive at the correct outputs.

We use the binary cross-entropy loss function. Layers are first trained using the Adam optimizer \cite{b18} for 30 epochs with a learning rate of 1e-3. We then continue to train the model using the Stochastic Gradient Descent (SGD) optimizer for gradient descent with a learning rate of 1e-4 for 50 epochs. SGD enables us to escape local minima of the loss function by using small, random movements. This process makes it easier for the CNN to find the global minimum of the loss function.

\subsubsection{Ensemble Multiple Models}
We use the process of ensemble averaging to combine our three separate models into one. This approach is superior to generating only one classifier as the various errors among each model due to overfitting or underfitting will average out, resulting in higher overall scores. This ensemble is called Deepwound.

When a new image is fed into Deepwound, it is independently delegated to each member WoundNet CNN for classification. The results from each algorithm are consolidated into one result matrix through majority-voting for the presence or absence of each label.

\section{Mobile Application Pipeline}
With a predicted 6.8 billion smartphones in the world by 2022 \cite{b19} mobile health monitoring platforms can be leveraged to provide the right care at the right time. In this research, we have developed a comprehensive mobile application, Theia, as a way to deliver our Deepwound model to patients and providers. Screenshots from the final app are shown in Figure 8.

Theia is a proof-of-concept of how Deepwound can assist physicians and patients in postoperative wound surveillance. The first component of the app is the "Quick Test." Physicians or patients can quickly photograph a wound and generate a wound assessment. A wound assessment provides positive or negative values for each label affiliated with a wound.

The other component of the app is the ability for a patient to track wounds over a period of time. Every day, the patient receives a notification to complete a daily wound assessment, where he/she provides an image of the surgical site, their current weight, and rate their pain in that area. This data is accumulated over a period of 30 days. Furthermore, patients can track many different variables that can affect their wound recovery such as medicine intake, the changing of their wound dressing, weight, and pain level. All of this information is charted out over time and can be converted into a PDF that can be sent to physicians or family. Finally, we provide easy access for the patient to directly contact their surgeon through the app itself.

With permission from the patient, this app can also be used to collect wound images to add to our dataset. The enlarged dataset can be used to further improve our deep learning algorithms. As more patients and surgeons use the app, more image data can be collected. This newly accumulated data can be used to train our CNNs even further, leading to a virtuous cycle of improving accuracy.

Deepwound is used to classify every image the user takes. The image is stored securely within the app and anonymized when sent to the server for image processing and classification via a multipart HTTP request.

\section{Results and Discussion}
We now present results for our computational model. We evaluate our CNN ensemble by calculating a variety of classification metrics (e.g accuracy, sensitivity, specificity, and F1 score), analyzing receiver operating characteristic curves, and generating saliency maps. This process is diagrammed in Figure 9.

\begin{figure}[H]
\centerline{\includegraphics[width=0.45\textwidth]{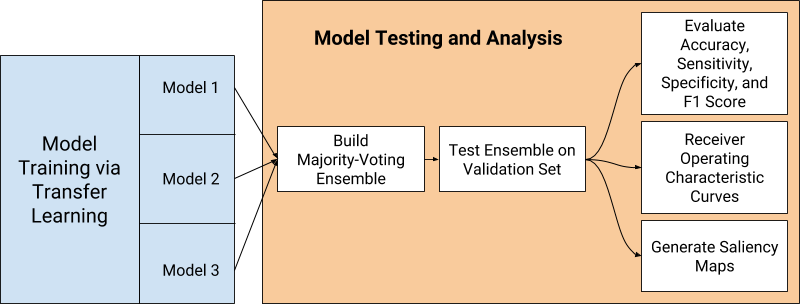}}
\caption{Model testing and analysis.}
\label{fig}
\end{figure}

\subsection{Classification Metrics}
We use a few different metrics to evaluate the performance of the ensemble as a whole. We use accuracy, sensitivity, specificity, and F1 Score (see Equation 1, Equation 2, Equation 3, Equation 4). The latter three are more reliable metrics than accuracy for this paper as they take into account the real effectiveness of the model at discerning the presence and absence of a particular ailment. Table II displays all of our scores.

\begin{equation}
accuracy = \frac{all\;correct}{all\;samples} \label{eq}
\end{equation}

\begin{equation}
sensitivity = \frac{true\;positive}{all\;positive}\label{eq}
\end{equation}

\begin{equation}
specificity = \frac{true\;negative}{all\;negative}\label{eq}
\end{equation}

\begin{equation}
F1\;score = 2 * \frac{sensitivity * specificity}{sensitivity + specificity}\label{eq}
\end{equation}

\begin{table}[H]
\caption{Reported classification metrics for the ensemble.}
\begin{center}
\begin{tabular}{|l|c|c|c|c|}
\hline
\textbf{Labels}    & \textbf{Accuracy} & \textbf{Sensitivity} & \multicolumn{1}{l|}{\textbf{Specificity}} & \multicolumn{1}{l|}{\textbf{F1 Score}} \\ \hline
Wound              & 0.82              & 0.93                 & 0.75                                      & 0.83                                   \\ \hline
Infection (SSI)    & 0.84              & 0.70                 & 0.70                                      & 0.70                                   \\ \hline
Granulation Tissue & 0.85              & 0.92                 & 0.73                                      & 0.81                                   \\ \hline
Fibrinous Exudate  & 0.83              & 0.71                 & 0.74                                      & 0.72                                   \\ \hline
Open Wound         & 0.83              & 0.96                 & 0.75                                      & 0.84                                   \\ \hline
Drainage           & 0.72              & 0.98                 & 0.55                                      & 0.70                                   \\ \hline
Steri Strips       & 0.97              & 0.88                 & 0.82                                      & 0.85                                   \\ \hline
Staples            & 0.95              & 0.83                 & 0.60                                      & 0.70                                   \\ \hline
Sutures            & 0.85              & 0.91                 & 0.57                                      & 0.70                                   \\ \hline
\end{tabular}
\label{tab1}
\end{center}
\end{table}

\subsection{Receiver Operating Characteristic and Area Under Curve}
The area under the curve (AUC) of a receiver operating characteristic (ROC) curve is a useful metric in determining the performance of a binary classifier. ROC curves graphically represent the trade-off at every possible cutoff between sensitivity and specificity. Better classifiers have higher AUC values for their ROC curves while worse classifiers have lower AUC values. We chart ROC curves and calculate AUC values for each possible label for an image, as seen in Figure 10.

\begin{figure}[H]
\centerline{\includegraphics[width=0.50\textwidth]{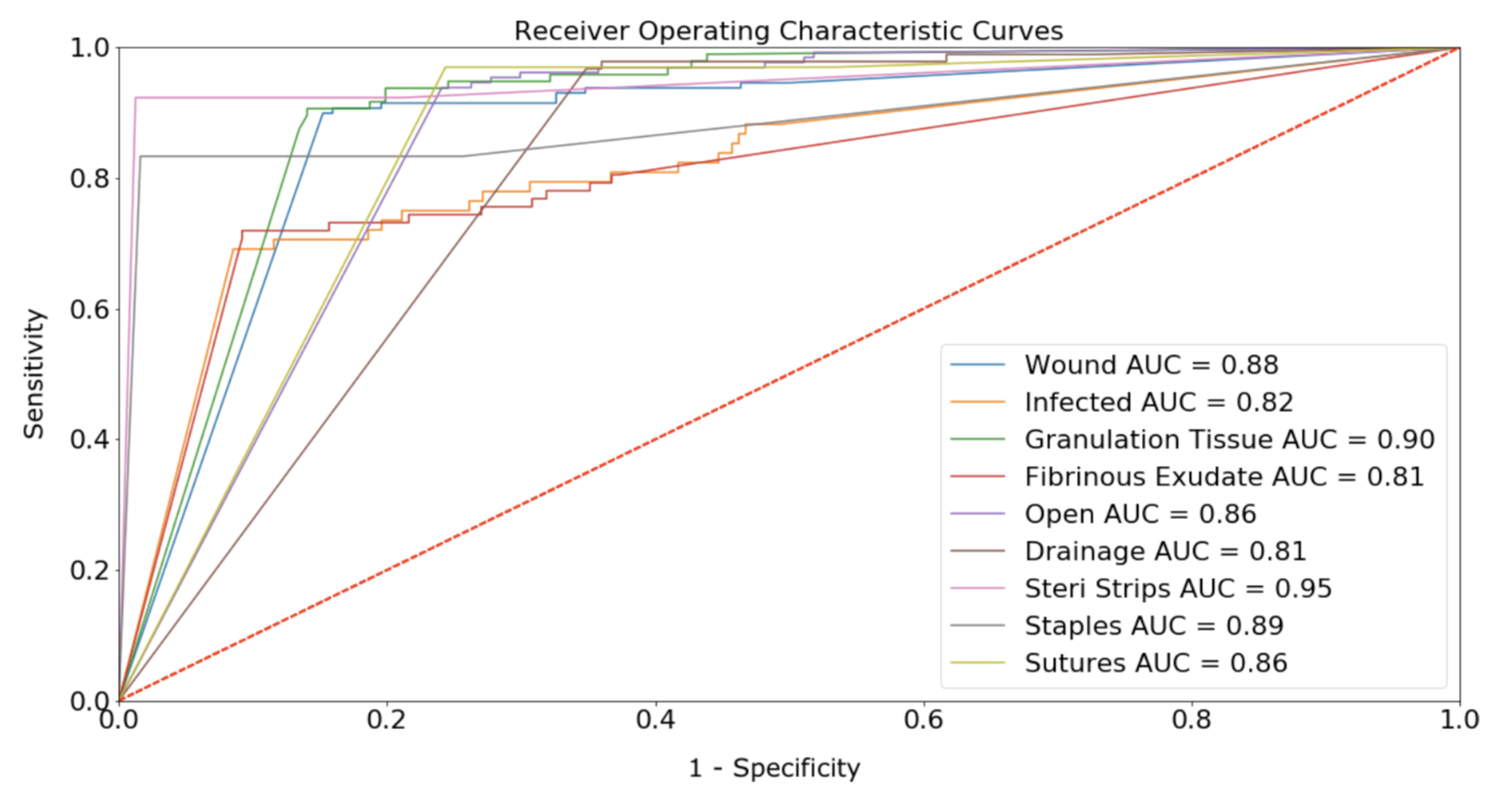}}
\caption{ROC curves for every class and their respective AUCs.}
\label{fig}
\end{figure}

\subsection{Saliency Maps}
When analyzing digital images using machine learning, it is important to understand why a certain classifier works. Saliency maps have been shown in the past as a way to visualize the inner workings of CNNs in the form of a heat map which highlights the features within the image that the classifier is focused on \cite{b20}. We generate saliency maps from one of CNNs on a couple of images in the validation set to ensure that our classifiers are identifying the regions of interest for a particular label in an image (see Figure 11). We can confirm that the attention of the model is drawn to the correct regions in the images.

\begin{figure}[H]
\centerline{\includegraphics[width=0.50\textwidth]{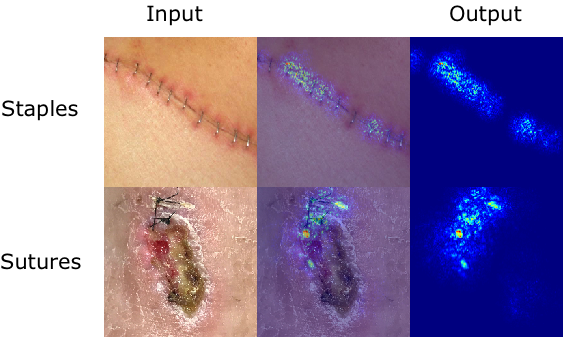}}
\caption{Judging CNN performance through saliency visualization.}
\label{fig}
\end{figure}

\section{Conclusions}
In summary, our work describes a new machine learning based approach using CNNs to analyze an image of a wound and document its wellness. Our implementation achieves scores that improve upon prior work by Wang et al. and Sanger et al. for F1 score and ROC AUC.

We acknowledge that our data set size is small and has some imbalance. This is a common problem in medical research as the data needs to be gathered over a sustained period of time with health compliant processes. We overcome these hurdles through the use of aggressive data augmentation, transfer learning, and an ensemble of three CNNs.

Our approach for analysis and delivery with a smartphone is a unique contribution. It enables several key benefits: tracking a patient remotely, ease of communication with the medical team and an ability to detect the early onset of infection. Wide spread use of such means can also enable automated data collection and classification at a lower cost, which in turn can improve the machine learning algorithm to be improved through re-training with a larger data set. Our mobile app can also generate comprehensive wound reports that can be used for the purpose of billing insurers, thus saving surgeons time.

\section{Future Work}
There are many ways to improve our algorithm. On a larger scale, it is necessary to gather more images for both training and testing. Creating a robust corpus of images will enable us to improve the performance of our method. More labeled images always lead to higher performances in the field of deep learning.

We would like to consider blur detection prior to analyzing our image. If the image is too blurry, we can send a message back to the user, requesting a clearer picture. There are many well-known techniques to accurately measure blur within an image.

We would also like to look into embedding our model into mobile devices directly without the need for a server. This will drastically increase speed for users and enable them to use the app in locations without access to the internet. Finally, we would like to extend our wound assessment framework by developing a computational model to track the healing of a wound using a time-series of images which can be collected using the current version of the mobile app.

\end{document}